\documentclass[10pt, a4paper]{article}
\usepackage{lrec2022} 
\usepackage{multibib}
\newcites{languageresource}{Language Resources}
\usepackage{graphicx}
\usepackage{tabularx}
\usepackage{soul}
\usepackage{titlesec}
\titleformat{\section}{\normalfont\large\bfseries\center}{\thesection.}{1em}{}
\titleformat{\subsection}{\normalfont\SmallTitleFont\bfseries\raggedright}{\thesubsection.}{1em}{}
\titleformat{\subsubsection}{\normalfont\normalsize\bfseries\raggedright}{\thesubsubsection.}{1em}{}
\renewcommand\thesection{\arabic{section}}
\renewcommand\thesubsection{\thesection.\arabic{subsection}}
\renewcommand\thesubsubsection{\thesubsection.\arabic{subsubsection}}

\usepackage{epstopdf}

\usepackage[colorlinks=true,linkcolor=black,citecolor=black,filecolor=black,urlcolor=black]{hyperref}
\usepackage{xstring}

\usepackage{color}
\usepackage{nert}
\usepackage{tikz-dependency}
\usepackage{adjustbox}

\usepackage{natbib}

\usepackage{leipzig}
\usepackage{gb4e}
\noautomath

\newcommand{\bio}[1]{\textsc{\textbf{\texttt{\color{red}{#1}}}}}

\title{MASALA: Modelling and Analysing the Semantics of Adpositions in Linguistic Annotation of Hindi}

\name{Aryaman Arora, Nitin Venkateswaran, Nathan Schneider} 

\address{Georgetown University \\
         Washington, D.C., USA\\
         \{\emldisplay{aa2190@georgetown.edu}{aa2190}, \emldisplay{nv214@georgetown.edu}{nv214}, \emldisplay{nathan.schneider@georgetown.edu}{nathan.schneider}\}\texttt{@georgetown.edu}}

\abstract{
We present a completed, publicly available corpus of annotated semantic relations of adpositions and case markers in Hindi. We used the multilingual SNACS annotation scheme, which has been applied to a variety of typologically diverse languages. Building on past work examining linguistic problems in SNACS annotation, we use language models to attempt automatic labelling of SNACS supersenses in Hindi and achieve results competitive with past work on English. We look towards upstream applications in semantic role labelling and extension to related languages such as Gujarati.
 \\ \newline  }

\begin{document}

\maketitleabstract

\section{Introduction}
Case markers express semantic roles, describing the relationship between the arguments they apply to and the action of a verb. Adpositions (prepositions, postpositions, and circumpositions) further express a range of semantic relations, including space, time, possession, properties, and comparison.

Languages have different strategies for encoding these kinds of semantic relations. Hindi--Urdu\footnote{Hindi and Urdu are two registers written in two different scripts of a single language (usually called `Hindi--Urdu' or `Hindustani') with a largely identical grammar. While our corpus is in Hindi in the Devanagari script, the linguistic portions of our work (e.g.~annotation guidelines) are applicable to Urdu as well.} uses a case-marking system along with a large postposition inventory \citep{kachru2006hindi,koul}. Idiosyncratic bundling of case and adpositional relations poses problems in many natural language processing tasks for Hindi, such as machine translation (\citealt{Ratnam2018}, \citealt{Jha2017}, \citealt{ramanathan-etal-2009-case}, \citealt{Rao1998}) and semantic role labelling (\citealt{pal-sharma-2019-dataset}, \citealt{gupta2019semantic}). Many models for these tasks rely on human-annotated corpora for training data, such as the one created for the Hindi--Urdu PropBank \citep{bhatt-etal-2009-multi}, and in \citet{kumar2019case}. The study of adposition and case semantics in corpora is also useful from a comparative\slash typological linguistic perspective, in comparing and categorizing the encoding of such relations across languages.

To that end, we release a completed Hindi corpus annotated for adposition and case semantic labels using the SNACS formalism \citep{schneider-18,snacs-guidelines}. We approach the problem of automatic tagging of these labels using a variety of language models and explore what these models learn. Drawing on parallel SNACS corpora in English, German, Mandarin, and Korean, we compare strategies for encoding semantic roles across languages.

\section{Background}
Hindi is a language of India, of the Indo-Aryan branch of the Indo-European family, and one of the best-resourced South Asian languages for research in natural language processing and computational linguistics \citep{joshi-etal-2020-state}. Hindi has a small number of core case markers as well as a large class of adpositions for signalling semantic relations. We will discuss the linguistic features of case and adposition in Hindi below, related work from linguistics in this area, and introduce the SNACS schema.

\subsection{Case and adposition in Hindi}
Hindi is generally described as having three layers of case/adposition: the three basic morphological cases (\cref{ex:morph}), a small class of case markers/clitics that indicate core arguments to verbs (\cref{ex:cases}), and a larger class of postpositions governed by the genitive \textit{kā} or ablative \textit{se} (\cref{ex:posts}) \citep{kachru2006hindi}.
\begin{exe}
    \ex \begin{xlist}
    \ex \label{ex:morph}bacc\p{e} `children', bacc\p{o\.{m}} `children.\Obl', bacc\p{o} `children.\Voc'
    \ex \label{ex:cases}us\p{ne} `she.\Erg', us\p{ko} `she.\Acc/\Dat'
    \ex \label{ex:posts} us\p{ke\_liye} `for her', us\p{ke\_nazdīk} `near her', us\p{ke\_\textit{under}} `under her' [code-switching]
    \end{xlist}
\end{exe}
\citet{masica1993indo} grouped these three ``layers'' on the basis of historical development. Diachronically, morphological cases are the remnants of the Indo-European case system (via Sanskrit) that largely encode syntactic information, the case markers are Middle Indo-Aryan developments from spatial adverbs (e.g.~Sanskrit \textit{upari} `above' > Hindi \textit{par} `\Loc-on') that encode fundamental semantic roles on complements, and postpositions are more recent developments that even include borrowings from Persian, Arabic, and English and which indicate more concrete, e.g.~spatial, relations between nominals.

The case markers most commonly mark relations between verbs and their arguments and adjuncts, followed by relations between nominals. Case markers in Hindi are highly multi-functional even when using coarse descriptors from linguistic typology; e.g.~\textit{se} is described as indicating the ablative, instrumental, comitative, or comparative cases depending on context, respectively exemplified in \cref{ex:se}.
\begin{exe}
    \ex \label{ex:se}\begin{xlist}
    \ex yahā\.{m} \p{se} jāō \hfill{} `Go away \p{from} here.'
    \ex cammac \p{se} k\textsuperscript{h}ānā \hfill{} `eating \p{with} a spoon'
    \ex us\p{se} milū\.{m}gā \hfill{} `I will meet \p{with} him.'
    \ex das \p{se} kam \hfill{} `less \p{than} ten'\end{xlist}
\end{exe}
That is not to say that \textit{se} is three different case markers; the semantic role of a \textit{se}-marked argument is just licensed by the predicate or other governor of the argument. Understanding how and in what context such markers indicate what semantic relations is an interesting problem. Thus far, there is no semantically-annotated corpus of case and postposition semantics in Hindi, which motivated our annotation of this corpus.

\subsection{Related work}
There is a great deal of work on case and adpositions in Hindi. In syntax, some research topics are syntactic differences between morphological case, case markers, and adpositions \citep{spencer2005case}, the issue of differential case marking in the ergative and dative--accusative \citep{bhatt1996object,DEHOOP2005321,de2009ergative,montrul2015differential,montaut2018rise}, word order \citep{mohanan1994case}, and agreement \citep{montrul2012erosion}.

On the other hand, there has been less research on the semantics of case and adpositions in Hindi. The mapping of case-marked arguments to lexical-semantic roles has been done in various computational projects \citep{begum-etal-2008-developing,vaidya-etal-2011-analysis}. \citet{paul-etal-2010-syntactic} is an investigation of paraphrasing nominal compound relations with case in Hindi and English.

\subsection{SNACS}\label{sec:snacs}
The Semantic Network of Adposition and Case Supersenses (SNACS; \citealp{schneider-18,snacs-guidelines}) is a multilingual annotation scheme with 50 supersenses that characterize the use of adpositions and case markers at a coarse level of granularity. This scheme is akin to linguistic models of argument structure such as semantic roles and theta roles (including traditional categories such as \psst{Agent} and \psst{Theme}), but expanded to include roles for adpositional relations, such as \psst{Whole} for whole--part, \psst{SocialRel} for interpersonal relations, etc. 

A useful feature of SNACS is the \textit{construal system} \citep{hwang-etal-2017-double}, which allows an annotator to give one label for the morphosyntactic role or inherent lexical meaning (\textbf{function}) and another label for the predicate-licensed semantic relation (\textbf{scene role}) of a token. This is expressed as \rf{SceneRole}{Function} if they differ. Examples of SNACS annotation for Hindi are given below.

\begin{exe}
\small
\ex \gll vah g\textsuperscript{h}ar \p{ke\_pās}\textsubscript{\psst{Locus}} hai\\
         {\Third\Sg} {home} {near} {\Cop.\Ind.\Third\Sg} \\
    \glt `He is near the house.'
\ex \gll mai{\. m} us \p{ko}\textsubscript{\psst{Theme}} k\textsuperscript{h}ā-tā hū{\. m}\\
         {\First\Sg} {\Third\Sg} {\Acc} {eat-\Ipfv.\M.\Sg} {\Cop.\Ind.\First\Sg}\\
    \glt `I eat that.'
\ex \gll mai{\. m} \p{ne}\textsubscript{\rf{Experiencer}{Agent}} nadī \p{ke\_pār}\textsubscript{\rf{Locus}{Path}} ek baccā dekh-ā\\
      {\textsc{1sg}} {\textsc{erg}} {river} {across} {one} {child.\textsc{nom}} {see-\textsc{pfv.m.sg}}\\
    \glt `I saw a child across the river.'
\end{exe}

SNACS, thus far, has been used to annotate the English STREUSLE corpus \citep{schneider-smith-2015-corpus}, \textit{The Little Prince} in English and translations of it into Korean \citep{hwang-etal-2020-k}, Mandarin \citep{peng-etal-2020-corpus}, and German \citep{german}. There has also been annotation of L2 English \citep{kranzlein-etal-2020-pastrie}. This effort has been accompanied by the release of guidelines for annotator training, including for English \citep{snacs-guidelines} and Hindi--Urdu \citep{hindi-snacs-guidelines}. Some earlier works also discussed linguistic issues in Hindi annotation \citep{arora-etal-2021-snacs}.

There is also an online interface for exploring SNACS corpora and interactive annotation guidelines: \url{http://www.xposition.org/} \citep{xposition}.

\section{Corpus and annotation}

\begin{table}
    \small
    \centering
    \begin{tabular}{lrrr}
        \toprule
        & \textbf{Count} & \textbf{\%} & \textbf{Types} \\
        \midrule
        \textsc{Chapters} & 27 \\
        \textsc{Sentences} & 1,580 \\
        \textsc{Tokens} & 16,882 \\
        \midrule
        \textsc{Targets} & 2,970 & & 70 \\
        \quad Case & 2,142 & 72.1\% & 7 \\
        \quad Emphatic & 382 & 12.9\% & 3 \\
        \quad Adpositions & 446 & 15.0\% & 60 \\
        \midrule
        \textsc{Construals} & 2,970 & & 136 \\
        \quad Role $=$ Fxn. & 1,886 & 63.4\% & 38 \\
        \quad Role $\neq$ Fxn. & 1,084 & 36.6\% & 98 \\
        \bottomrule
    \end{tabular}
    \caption{\label{table:statistics} Cumulative statistics of the Hindi corpus.}
\end{table}

The corpus was the entirety of \textit{Nanhā Rājkumār}, the Hindi translation of the \textit{The Little Prince} by Antoine de Saint-Exupéry.\footnote{The corpus is available at \url{https://github.com/aryamanarora/carmls-hi}.} We used the SNACS annotation scheme, of which a brief overview is given in \cref{sec:snacs}. Annotation was done by two Hindi speakers: A (the first author, who is a native speaker) and B (the second author, who is highly proficient) during June 2020--January 2021, and annotation guidelines were developed simultaneously \citep{hindi-snacs-guidelines}. \Cref{table:statistics} contains statistics about the final corpus, which was released in CoNLL-U-Lex format with Universal Dependencies annotations generated with Stanza \citep{qi-etal-2020-stanza}.

There were two phases of annotation. In the first, A annotated the whole corpus (including all case markers and adpositions) and developed basic guidelines. In the second, B annotated chapter-by-chapter and A and B adjudicated disagreements concurrently. B also annotated focus markers, which were not included as targets in the first phase. A final pass was then conducted over the whole corpus to reconcile any remaining annotation disagreements.

\subsection{Annotation targets}
\begin{table*}[t]
    \small
    \centering
    \begin{tabular}{@{}clr}
        & \textit{Type} & \textit{\%} \\
        \cmidrule(lr){2-3}
      \multirow{7}{*}{\rotatebox{90}{\textbf{Case Markers}}} &
        \textit{k\={a}} (\Gen) & 24.3 \\
       & \textit{ko} (\Acc/\Dat) \vphantom{X} & 15.8 \\
       & \textit{ne} (\Erg) & 10.2 \\
       & \textit{se} (\Ins/\Abl/\Com) & 9.5 \\
       & \textit{me\.{m}} (\Loc-in) & 6.3 \\
       & \textit{par} (\Loc-on) & 5.2 \\
       & \textit{tak} (\All) & 0.8 \\\\
        \cmidrule(lr){2-3}
       \multirow{3}{*}{\rotatebox{90}{\textbf{Focus}}} & \textit{to} (contrastive) & 6.2\\
       & \textit{hī} (``even'') & 3.6 \\
       & \textit{bhī} (``also'') & 3.0 \\\\
        \cmidrule(lr){2-3}
       \multirow{5}{*}{\rotatebox{90}{\textbf{Adpositions}}} & \textit{ke lie} (``for'') & 3.3\\
       & \textit{ke p\={a}s} (``near'') & 1.0 \\
       & \textit{s\={a}} (``-ish'') & 1.0 \\
       & \textit{k\={\i} tarah} (``like'') & 1.0 \\
       & \textit{jaise} (``like'') & 1.0 \\
    \end{tabular}\hfill
    \begin{tabular}{lr}
        \textit{Scene role} & \textit{\%} \\
        \cmidrule(lr){1-2}
        \psst{Experiencer} & 8.8 \\
        \psst{Originator} & 6.8 \\
        \psst{Theme} & 6.5 \\
        \psst{Locus} & 5.2 \\
        \psst{Topic} & 5.1 \\
        \psst{Gestalt} & 4.7 \\
        \psst{Agent} & 4.7 \\\\
        \cmidrule(lr){1-2}
        \psst{Focus} & 9.6 \\
        \texttt{\textasciigrave d} & 2.9 \\
        \texttt{NONSNACS} & 0.4 \\\\
        \cmidrule(lr){1-2}
        \psst{CompRef.} & 2.1 \\
        \psst{Purpose} & 1.1 \\
        \psst{Explanation} & 1.1 \\
        \psst{Time} & 1.0 \\
        \psst{Extent} & 0.9 \\
    \end{tabular}\hfill
    \begin{tabular}{lr}
        \textit{Function} & \textit{\%} \\
        \cmidrule(lr){1-2}
        \psst{Agent} & 11.0 \\
        \psst{Theme} & 10.2 \\
        \psst{Gestalt} & 9.9 \\
        \psst{Recipient} & 7.7 \\
        \psst{Locus} & 6.5 \\
        \psst{Source} & 4.0 \\
        \psst{Topic} & 3.1 \\\\
        \cmidrule(lr){1-2}
        \psst{Focus} & 9.6 \\
        \texttt{\textasciigrave d} & 2.9 \\
        \texttt{NONSNACS} & 0.4 \\\\
        \cmidrule(lr){1-2}
        \psst{CompRef.} & 2.8 \\
        \psst{Locus} & 1.7 \\
        \psst{Beneficiary} & 1.3 \\
        \psst{Purpose} & 1.1 \\
        \psst{Explanation} & 1.1 \\
    \end{tabular}\hfill
    \begin{tabular}{r@{$\leadsto$}lr}
        \textit{Scene role} & \textit{Function} & \textit{\%} \\
        \cmidrule(lr){1-3}
        \psst{Theme} & \psst{Theme} & 5.7 \\
        \psst{Experiencer} & \psst{Recipient} & 5.2 \\
        \psst{Originator} & \psst{Agent} & 4.9 \\
        \psst{Gestalt} & \psst{Gestalt} & 4.5 \\
        \psst{Locus} & \psst{Locus} & 4.0 \\
        \psst{Agent} & \psst{Agent} & 3.4 \\
        \psst{Topic} & \psst{Topic} & 3.0 \\\multicolumn{1}{l}{}\\
        \cmidrule(lr){1-3}
        \psst{Focus} & \psst{Focus} & 9.6 \\
        \texttt{\textasciigrave d} & \texttt{\textasciigrave d} & 2.9 \\
        \texttt{NONSNACS} & \texttt{NONSNACS} & 0.4 \\\multicolumn{1}{l}{}\\
        \cmidrule(lr){1-3}
        \psst{CompRef.} & \psst{CompRef.} & 2.1 \\
        \psst{Purpose} & \psst{Purpose} & 1.1 \\
        \psst{Expl.} & \psst{Expl.} & 1.1 \\
        \psst{Extent} & \psst{Extent} & 0.9 \\
        \psst{Experiencer} & \psst{Benef.} & 0.9 \\
    \end{tabular}
    \caption{\label{tab:roles} Breakdown of label counts along various dimensions, divided between case markers and adpositions. \textbf{Each of the 8 tables is independent.} (E.g., the topmost `Scene role' table shows that 8.8\% of annotated targets in the corpus are case markers with the scene role \psst{Experiencer}.)}
\end{table*}

Following \citeposs{masica1993indo} analysis of Indo-Aryan languages, we annotated the Layer II and III function markers in Hindi. These include all of the simple case markers\footnote{\textit{ne} (ergative), \textit{ko} (dative-accusative), \textit{se} (instrumental-ablative-comitative), \textit{k\={a}}/\textit{ke}/\textit{k\={\i}} (genitive), \textit{me\.{m}} (locative-IN), \textit{tak} (allative), \textit{par} (locative-ON). Declined forms of the pronouns (including the reflexive \textit{apn\={a}}) were also included.} and all of the adpositions.

We also decided to annotate the suffix \textit{v\={a}l\={a}} when used in an adjectival sense (e.g. \textit{choṭā-vālā kamrā} `the room that is small'), the comparison terms \textit{jais\={a}} and \textit{jaise}, the extent and similarity particle \textit{sā} (\textit{choṭā-sā kamrā} `small-ish room'), and the emphatic particles \textit{bhī}, \textit{hī}, \textit{to} \cite[137--156]{koul}. All of these modify the preceding token and mediate a semantic relation between their object and the object's governor, just as conventionally-designated postpositions do.

The directly-declined Layer I cases of nominative, oblique, and vocative were not annotated due to the much greater annotation load that would involve and how much greater the breadth of the annotations would be relative to other SNACS-annotated languages. This means verbal arguments without case clitics were not annotated. However, future work (especially with application to semantic role labelling) would benefit from such annotations, and similar work has been done on SNACS annotation of non-adpositionally-marked subjects and objects in English \citep{shalev-etal-2019-preparing}.

\subsection{Linguistic issues}

Several linguistic features of Hindi--Urdu adposition and case semantics posed difficulties in annotating. Some are examined below. The annotation process itself relied on grammatical analyses of Hindi such as \citet{koul}, dictionaries \citep{mcgregor,dasa}, and native speaker judgements.

\paragraph{Functions for case markers} Case markers encode little lexical content relative to adpositions. \Cref{tab:roles} shows the dominance of case markers in every category; given their versatility, delineating their prototypical functions is difficult. For example, a comparative in Hindi--Urdu is expressed with the ablative case marker \textit{se}---should the function be \psst{Source} (as expected for the ablative case) or the narrower \psst{ComparisonRef} in this sense? This is an unresolved question; in labelling, we chose narrower functions when their use seemed to be a relation that is not completely supplied by the predicate.

In other cases, with highly polysemous markers such as \textit{se}, it is difficult to pick a single function corresponding to an obvious grammatical case. For example, the verb \textit{pūchnā} `to ask' takes an argument, marked with \textit{se}, indicating the person being asked. This instance of \textit{se} could be construed as the ablative case (reflecting the return of a response from the person asked) or the comitative case (indicating a co-participant in communication, exactly as for verbs such as \textit{kahnā} `to say').

\begin{exe}
\ex \gll us-\p{se} apnā savāl pūcho.\\
         {\Third\Sg.\Obl-?} {self.\Gen} {question} {ask.\Imp}\\
    \glt `Ask them:\rf{Recipient}{?} your question.'
\end{exe}
To resolve this issue we looked to typological evidence, in keeping with SNACS's multilingual aims: the closely-related language Punjabi, which has separate ablative (\textit{to\.{m}}) and comitative (\textit{nāl}) markers, uses the ablative in this construction, so we labelled the function \psst{Source}.

\paragraph{Non-nominative/ergative subjects} The \psst{Agent} is prototypically expressed with the ergative case marker \textit{ne} or the unmarked nominative. To express modality, Hindi--Urdu, like other Indo-Aryan languages, employs various aspectual light verbs along with differential subject marking \citep{DEHOOP2005321}. One example is the dative subject indicating obligation:

\begin{exe}
\ex
\begin{xlist}
\ex \gll mai\.{m}-\p{ne} likh\={a}\\
         {\First\Sg-\Erg} write.\Prf\\
    \glt `I:\rf{Originator}{Agent} wrote it.'
\ex \gll mujh-\p{ko} likhn\={a} pa\d{r}\={a}\\
        {\First\Sg.\Obl-\Dat} {write.\Inf} {fall.\Prf}\\
    \glt `I:\rf{Originator}{?} had to write it.'
    \label{obligation}
\end{xlist}
\end{exe}

In these, the subject's scene role is \psst{Originator} as it is a producer of writing. In \cref{obligation}, an expression of obligation, the subject is not only compelled to act by some outer force (fitting a \psst{Theme}) but is also performing the action unaided (\psst{Agent}). SNACS currently cannot resolve the conflict between these two equally valid functions; we currently label \cref{obligation} as \rf{Originator}{Recipient} in keeping with the morphosyntax of the dative subject. The issue is a broader problem of dealing with force dynamics in semantic role labelling, and may require new labels.

Other unconventional subjects are less problematic. South Asian languages near-universally have dative subject \psst{Experiencer}s \citep{verma1990experiencer}.\footnote{Some South Asian languages also have dative \psst{Possessor}s.} For these, the prototypical \psst{Recipient} subject is fitting. The passive subject also has the unambiguous function of \psst{Agent}, just as the English passive \p{by}.

\paragraph{Causative constructions} Indo-Aryan languages, through suffixation, derive indirect and direct causative verbs from intransitive verbs. Indirect causatives take an argument in the instrumental case that is an \textit{impelled agent}, grammatically distinguished from a true \psst{Instrument}:
\begin{exe}
\ex \gll us-ne c\={a}bh\={\i}=\p{se} darv\={a}z\={a} khol\={a}\\
         {\Third\Sg.\Erg} {key.\Obl=\Ins} {door.\Nom} open.\Prf\\
    \glt `She opened the door [with a key]:\psst{Instrument}.'
\ex \gll us-ne m\={a}lik=\p{se} darv\={a}z\={a} khulv\={a}y\={a}\\
         {\Third\Sg.\Erg} {owner.\Obl=\Ins} {door.\Nom} open.\Ind.\Caus.\Prf\\
    \glt `She made [the landlord]:\psst{?} open the door.'
\end{exe}
Much like an obligated agent, the impelled agent takes part in two events, exhibiting properties of both \psst{Agent} and \psst{Theme}. Furthermore, an impelled agent can control \psst{Instrument}s of its own, and there cannot be two participants in the scene with the same semantic role \citep{begum-sharma-2010-preliminary}. For SNACS, \citet{shalev-etal-2019-preparing} mentioned similar issues in English.

This construction was rare in our corpus, but we find the best solution for this is a new label for animate and ambiguously volitional counterparts to \psst{Instrument} in the SNACS hierarchy, much like the distinction between inanimate \psst{Causer} and animate \psst{Agent}.

\paragraph{Emphatic particles} Following work on SNACS for Korean, which created a new label \psst{Focus} for ``postpositions that indicate the focus of a sentence (FOC), contributing information such as contrastiveness, likelihood, or value judgements'' \citep{hwang-etal-2020-k}, we found that the Hindi emphatic particles \textit{hī} `only', \textit{bhī} `also, too', \textit{to} (contrastive), and some uses of \textit{tak} `even' function as focus postpositions and thus merited annotation.

\subsection{Corpus analysis}
\begin{table}[]
    \centering
    \small
    \begin{tabular}{lrrrr}
    \toprule
    \textbf{Target} & $n$ & \textbf{Scene} & \textbf{Fxn} & \textbf{Cons}  \\
    \midrule
\textit{ke bāre me\.{m}} (``about'') & 23 & 1.00 & 1.00 & 1.00 \\
\textit{ke lie} (``for'') & 95 & 0.88 & 0.96 & 0.87 \\
\textbf{\textit{ne}} (\Erg) & 288 & 0.89 & 0.98 & 0.87 \\
\textit{kī tarah} (``like'') & 29 & 0.83 & 0.97 & 0.83 \\
\textbf{\textit{ko}} (\Acc/\Dat) & 446 & 0.83 & 0.95 & 0.81 \\
\textbf{\textit{par}} (\Loc-on) & 107 & 0.83 & 0.86 & 0.79 \\
\textbf{\textit{me\.{m}}} (\Loc-in) & 180 & 0.80 & 0.86 & 0.77 \\
\textbf{\textit{se}} (\Ins/\Abl/\Com) & 253 & 0.79 & 0.81 & 0.68 \\
\textbf{\textit{kā}} (\Gen) & 682 & 0.72 & 0.79 & 0.66 \\
\textit{jaise} (``like'') & 28 & 0.57 & 0.86 & 0.54 \\
\textit{ke pās} (``near'') & 30 & 0.97 & 0.53 & 0.53 \\
\textit{vālā} (adjectival) & 22 & 0.36 & 0.41 & 0.36 \\
\textbf{\textit{tak}} (\All) & 23 & 0.65 & 0.43 & 0.35 \\
    \bottomrule
    \end{tabular}
    \caption{Raw agreement on targets with at least 20 doubly-annotated instances in the corpus, sorted by agreement on the construal. Case markers are in \textbf{bold}.}
    \label{tab:agreement}
\end{table}

\paragraph{Annotator agreement}
$2,368$ targets ($79.7\%$ of the total) were annotated independently by both annotators. In the first round of annotation by annotator A, the focus markers and a small number of case markers were not annotated. 

Cohen's $\kappa$ between both annotators for double-annotated targets was 0.78 on scene roles, 0.85 on functions, and 0.73 on construals (role$\leadsto$function), all of which are very high even compared to previous work on SNACS. It is not surprising that functions, which are inherent to the target type and less dependent on semantics, are easier to annotate than scene roles.

\Cref{tab:agreement} shows raw agreements on high-frequency targets, sorted by agreement on the construal label. Among the case markers, \textit{ne} (\Erg) and \textit{ko} (\Acc/\Dat) are the easiest to annotate, which is unsurprising given that their usage is very consistent syntactically (subjects and objects/indirect objects, respectively). The low agreement on \textit{tak} (\All, ``until, up to'') was due to uncertainty over whether it indicates the endpoint of movement (\psst{Goal}) or the length of the distance covered to the endpoint (\psst{Extent}); after adjudication, we standardised on the latter. The adposition \textit{ke pās} ``near'' had a similar problem, where we disagreed on whether \psst{Possessor} was an inherent syntactic function of it or a semantic extension of its spatial use.

\paragraph{Marker and tag distributions}
\begin{table}[]
    \small
    \centering
    \begin{tabular}{lrr}
        \toprule
        \textbf{Target} & $\widehat{H}$ & $n$ \\
        \midrule
\textbf{\textit{se}} (\Ins/\Abl/\Com) & 3.90 & 281 \\
\textbf{\textit{kā}} (\Gen) & 3.88 & 723 \\
\textbf{\textit{me\.{m}}} (\Loc-in) & 3.17 & 187 \\
\textbf{\textit{par}} (\Loc-on) & 2.78 & 155 \\
\textbf{\textit{ko}} (\Acc/\Dat) & 2.75 & 470 \\
\textit{ke lie} (``for'') & 2.48 & 97 \\
\textit{ke p\=as} (``near'') & 2.00 & 31 \\
\textit{jaise} (``like'') & 1.85 & 29 \\
\textit{v\=al\=a} (adjectival) & 1.83 & 28 \\
\textbf{\textit{tak}} (\All) & 1.79 & 24 \\
\textbf{\textit{ne}} (\Erg) & 1.64 & 302 \\
\textit{to} (contrastive) & 1.27 & 185 \\
\textit{k\={\i} tarah} (``like'') & 0.74 & 29 \\
\textit{s\=a} (``-ish'') & 0.47 & 31 \\
\textit{ke b\=are me\.m} (``about'') & 0.00 & 23 \\
\textit{bh\=\i} (``also'') & 0.00 & 90 \\
\textit{h\=\i} (``even'') & 0.00 & 107 \\
        \bottomrule
    \end{tabular}
    \caption{Estimated entropy of targets with at least 20 instances in the corpus. Case markers are in \textbf{bold}.}
    \label{tab:entropy}
\end{table}
Counts of targets and labels are presented in \cref{tab:roles}, which shows that case markers generally indicate core arguments of verbs (e.g.~\psst{Agent} as in subjects of verbs) and basic spatial relations (\psst{Locus}, \psst{Source}), focus markers have discourse uses, and adpositions indicate non-core adjuncts (e.g.~\psst{Purpose} `in order to').

Since Hindi has case markers, annotated targets were dominated by a few types with very large semantic breadth. We can operationalise a measure of \textbf{semantic range} using the entropy of the distribution of scene role labels, which are a coarse representation of semantics, for each case marker. Given a distribution $x$ (the scene roles) with classes $K$, Shannon entropy (in bits) is defined as:
\begin{equation}
    H(x) = -\sum_{k=1}^K p(x_k) \log_2 p(x_k)
\end{equation}
We further adjust for the sample size and distribution using the entropy estimator due to \citet{chao2003nonparametric}, which is suited for linguistic distributions \citep{entropy}. In \cref{tab:entropy}, we report entropy of scene role for adpositions and case markers with at least 20 occurrences in the corpus. Case markers, as expected, occupy the top 5 places. However, \textit{tak} (``until, up to'') and ergative-case \textit{ne} are much less semantically diverse than the other case markers. Some of the more frequent adpositions are also very semantically diverse, but most are not and form a long tail.

\section{Automatic tagging}

Given the recent abundance of language models (both multi- and monolingual) for Hindi, we were interested in how well SNACS labels could be automatically tagged. To that end, we trained a neural sequence tagger on the task of adposition and case marker segmentation and tagging of scene role and function. This is a subinstance of the \textbf{lexical semantic recognition} (LSR) task first proposed in \citet{liu-etal-2021-lexical}, who approached it with models similar to those used for named entity recognition (NER). Our tagger feeds the output of a contextual language model through a biLSTM then to a CRF which emits the final tagging. We loaded language models through HuggingFace \citep{wolf-etal-2020-transformers} and implemented our models with PyTorch \citep{torch} and AllenNLP \citep{gardner-etal-2018-allennlp}.

\begin{table}
\small
    \centering
    \begin{tabular}{llrr}
        \toprule
        \textbf{Language} & \textbf{Model} & \textbf{Scene} & \textbf{Fxn} 
        \\
        \midrule
        English & \citet{schneider-etal-2018-comprehensive} & 58.2 & 66.7 \\
        & \citet{liu-etal-2021-lexical} & \textbf{71.9} & \textbf{81.0}
        \\
        \midrule
        Hindi & Baseline & 40.1 & 56.2 
        \\
        & IndicBERT & 41.1 & 59.0 \\
        & mBERT & 52.0 & 68.7 \\
        & MuRIL & 55.4 & 70.6 \\
        & XLM-R & 58.7 & 74.3 \\
        & \textit{IT distilBERT} & 69.1 & 78.7 \\
        & \textit{IT BERT} & \textbf{71.4} & \textbf{81.8} \\
        \bottomrule
    \end{tabular}
    \caption{F1 scores on Hindi test set ($n=158$ sentences), only evaluated on gold and predicted \bio{B} tags, compared with past scores on English SNACS tagging on the STREUSLE corpus. Baseline scores are based on picking the most common tag for a given target. Language models in italics are monolingual.}
    \label{tab:results}
\end{table}

\subsection{Data preparation}
In preparation for training a classifier, we converted the SNACS labels to the \bio{BIO} scheme for sequence tagging. We only marked the label to be predicted on the \bio{B}-tag of each sequence. In the case of the \bio{B}-tagged-word being segmented into subwords by the language model being used, we labelled all non-initial subwords as \bio{I}.

For example, using multilingual BERT the phrase \textit{uske p\={\i}che} `behind them (sg.)' is tokenised into subwords and tagged for scene role labels as:

\begin{center}
    \begin{tabular}{cccc}
        \textit{\_us\textbf{ke}} & \textit{\textbf{\_p\={\i}}} & \textit{\textbf{cha}} & \textit{\textbf{e}} \\
        \bio{B}-Locus & \bio{I} & \bio{I} & \bio{I} \\
    \end{tabular}
\end{center}

The sentences in the dataset are randomly split 80/10/10 between train/dev/test. Training occurs on the train set with period checks against the development set to measure convergence. Scores are reported on the test set.

\subsection{Model}
The language models we tested are IndicBERT \citep{kakwani-etal-2020-indicnlpsuite}, the original multilingual BERT \citep{devlin-etal-2019-bert}, MuRIL \citep{khanuja2021muril}, XLM-RoBERTa \citep{conneau-etal-2020-unsupervised}, and some of the models from the Indic-Transformers library \citep{indictransformers}.

The outputs from the language models are inputted to a 2-layer biLSTM with dropout of 0.3. Its output goes to a CRF, and the highest probability tags are outputted through Viterbi decoding. The number of epochs trained $\{30, 60\}$, the learning rate $\{0.0001, 0.0002, 0.0005, 0.001\}$, and LSTM layer size $\{64, 128, 256, 512\}$ are manually tuned hyperparameters. We did experiment with other architectures (e.g.~RNNs, Transformers instead of the LSTM) but this was the best architecture we found.

\subsection{Results}

We report F1 scores on tagging in \cref{tab:results}. The best model is the Indic-Transformers BERT, with a distilled version (that is more efficient) coming in a close second. It is surprising that multilingual language models perform much worse than monolingual ones; IndicBERT, for example, is barely better than the baseline. These results are also competitive with F1 scores on English SNACS tagging, which bodes well for future work on multilingual SNACS given the complexity of the Hindi case marker system.

One issue that was prevalent across models was tokenisation errors involving the Devanagari script. The Indic-Transformers models droppped the vowel markers for \textit{u}, \textit{ū}, \textit{e}, \textit{ai}, the \textit{bindu} (nasalisation marker), and the \textit{halant} (vowel-killer) while tokenising. Somehow, they still were the highest-performing models; it is likely that with a fixed tokeniser and retraining they could have been even better.

Nevertheless, these are promising results, especially considering that the complex Hindi case system requires knowledge of verb frame semantics to accurately tag with SNACS.

\section{Conclusion}

We released an annotated corpus for Hindi of semantic relations encoded by case markers and adpositions, using the multilingual SNACS schema. We presented analysis of the distribution of labels and annotator agreement, explored linguistic issues encountered in annotation that pose problems for SNACS, and ran experiments on automatic sequence tagging for SNACS in Hindi with language models and biLSTM-CRF. We show that this is a feasible computational task and hope that this guides further work on SNACS for other languages, especially for those related to Hindi.

Future work on SNACS could consider multilingual comparisons, building upon work on aligning Korean and English annotations \citep{hwang-etal-2020-k} and multilingual tagging as explored in this paper. Particularly, there is ongoing work on SNACS annotation of Gujarati, a langauge closely related to Hindi; multilingual tagging of the two would be an interesting next task. Leveraging SNACS annotations for upstream tasks is also underexplored, despite a growing interesting in the semantic relations encoded in prepositions which have otherwise been understudied in NLP \citep{elazar2021textbased}. We hope that this corpus will also be useful for future study on semantics-reliant tasks in Hindi.

\section*{Acknowledgements}
We thank members of the CARMLS group (particularly Maitrey Mehta, Jena Hwang, and Vivek Srikumar) for stimulating discussions on case and adposition semantics, students in NERT, and the three anonymous reviewers for their helpful feedback.

\section{Bibliographical References}\label{reference}

\bibliographystyle{lrecnat}
\bibliography{lrec2022-example,anthology}


\end{document}